\pdfoutput=1
\documentclass{article}
\usepackage{preprint_style,times}

\usepackage{algorithm}
\usepackage{algpseudocode}

\usepackage{amsmath,amsfonts,bm}

\def\eqref#1{equation~\ref{#1}}

\def\1{\bm{1}}

\DeclareMathAlphabet{\mathsfit}{\encodingdefault}{\sfdefault}{m}{sl}
\SetMathAlphabet{\mathsfit}{bold}{\encodingdefault}{\sfdefault}{bx}{n}

\usepackage{hyperref}
\usepackage{url}
\usepackage{graphicx}
\usepackage{booktabs}
\usepackage{amssymb}
\usepackage{xcolor}
\usepackage{microtype}

\title{Do Latent World Models Really Select\\the Right Actions?}

\author{Zhengshu Zhang \\
University of Southern California \\
\texttt{zhengshu@usc.edu}
\And
Zhiyuan Li \\
Aalto University \\
\texttt{zhiyuan.li@aalto.fi}
}

\preprintfinalcopy
\begin{document}

\maketitle

\begin{abstract}
This paper shows that the planning success of a latent world model does not reveal how well its score selects actions.
Latent world models score candidate action sequences by the distance between their predicted embeddings and the embedding of a goal image, and their strong planning results are naturally read as evidence that this score picks good actions.
Our study is simple: we take two released models, JEPA-WM and DINO-WM, execute the action sequences their planners consider, and compare what the score prefers with what actually happens.
We make two observations.
First, the score is informative but unreliable: it picks better actions than chance, yet it often passes over a successful sequence that the search has already found.
Second, planning still succeeds, helped by a wider search and by replanning from new observations.
With the released search budget, PointMaze planning succeeds in 91\% of episodes, yet the top-scored candidate at the end of a search is no more likely to succeed than a random one.
These observations suggest that the score should be measured directly, next to the success rate.

\end{abstract}

\section{Introduction}
\label{sec:intro}

\begin{figure}[!t]
    \centering
    \includegraphics[width=\linewidth]{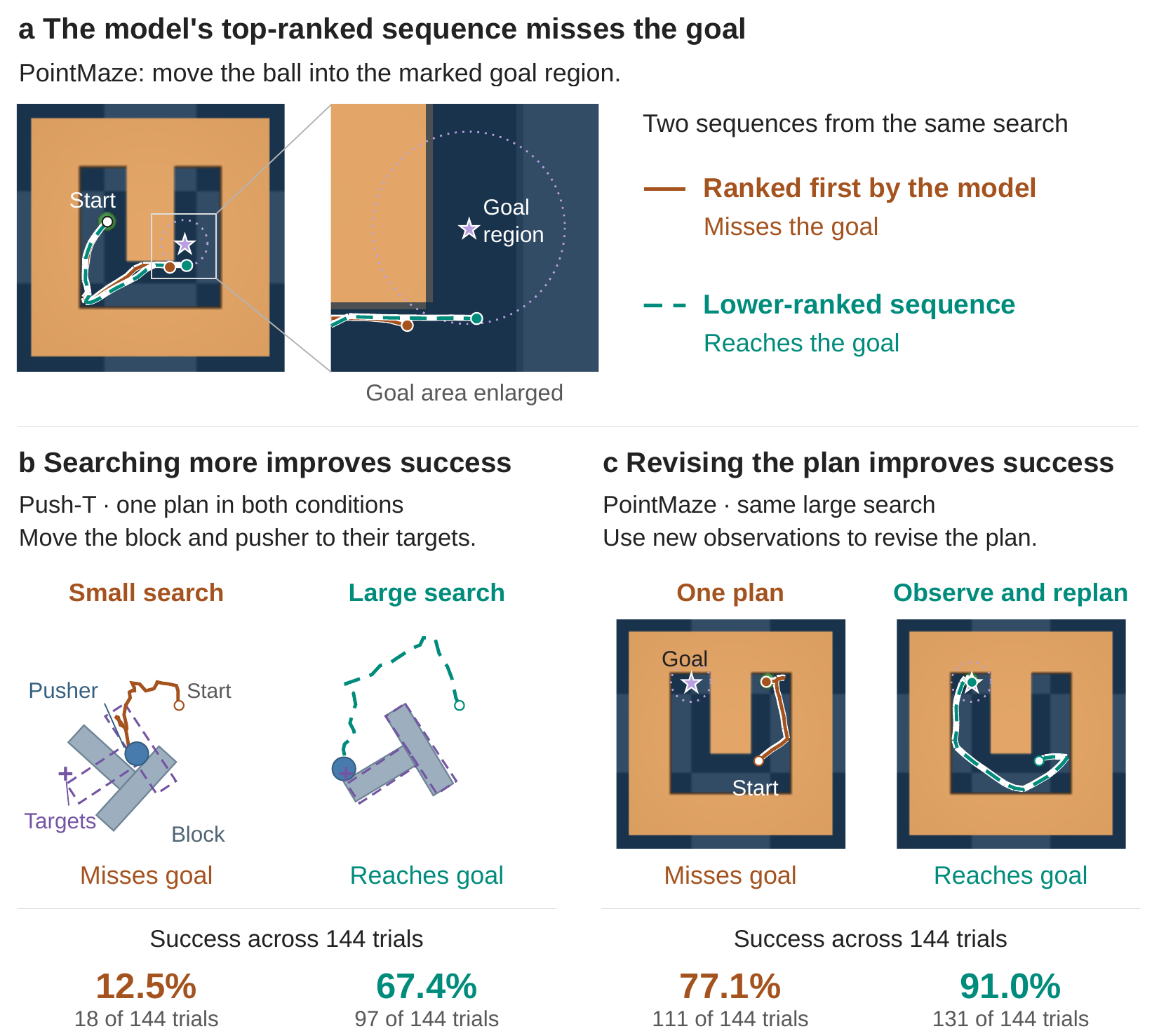}
    \caption{\textbf{Planning can succeed even when the model favors unsuccessful action sequences.} (a) In the PointMaze example, the model's top-ranked sequence misses the goal, while a lower-ranked sequence reaches it. (b) A wider search improves Push-T success even when the agent follows a single plan. (c) Replanning from new observations improves PointMaze success at the larger search budget. Purple marks the goal. The paths illustrate individual episodes; success rates summarize 144 paired trials.}
    \label{fig:teaser}
\end{figure}

World models help agents plan by predicting the consequences of their actions.
For visual tasks, latent world models encode images into learned representations and predict how these representations change under candidate action sequences~\citep{zhou2024dinowm}.
The planner scores each action sequence by how close its predicted outcome is to the goal image in representation space.
This allows the same learned model to support different goals, each specified by an image of the desired outcome~\citep{zhou2024dinowm}.
This approach has achieved strong results on navigation and manipulation tasks~\citep{zhou2024dinowm,terver2025jepawms}.

These results are commonly read as evidence that the score selects the right actions.
But a success rate is produced by the score together with the search that proposes candidate action sequences and the replanning that revises a plan from new observations.
Driven by this, a question arises: \emph{does the score of a latent world model really select the right actions?}

This paper answers the question by a direct measurement.
We take two released models, JEPA-WM and DINO-WM, in their own evaluation episodes, execute the action sequences that their planners consider, and compare what the score prefers with what actually happens.
We report that the score is informative but unreliable.
It picks better actions than chance on three of four simulated tasks, yet it often passes over a successful sequence that the search has already found (Figure~\ref{fig:teaser}a).
With a small search on PointMaze, a successful sequence is among the planner's final candidates in 90.3\% of episodes, and the top-scored one succeeds in 56.2\%.

We show that planning succeeds nonetheless, for both models and on every task we run in closed loop.
Based on these experiments, we observe the following:

(i) A wider search raises success with the same score.
On Push-T, a single plan succeeds in 12.5\% of episodes with a small search and in 67.4\% with the released one (Figure~\ref{fig:teaser}b).
After the larger search, the final candidates contain a successful sequence far more often.

(ii) Replanning raises success further.
The agent executes the first part of a plan, observes the result, and plans again.
On PointMaze, success rises from 77.1\% with a single plan to 91.0\% with replanning (Figure~\ref{fig:teaser}c).

(iii) After a wide search, the final choice of the score makes little difference.
The candidate scored highest succeeds about as often as one drawn at random from the final candidates, so a high success rate at this budget says little about how well the score selects.

These results challenge the reading of planning success as evidence that a latent world model selects the right actions.
The score helps the search, but a success rate alone cannot tell how well it selects.
We therefore suggest measuring the score directly, on executed candidates, next to the usual success rate.
Our study covers two released models and four simulated tasks, and we hope it provides useful evidence for evaluating world models that plan.

\section{Background and Related Work}
\label{sec:background}

\begin{figure}[!t]
    \centering
    \includegraphics[width=\linewidth]{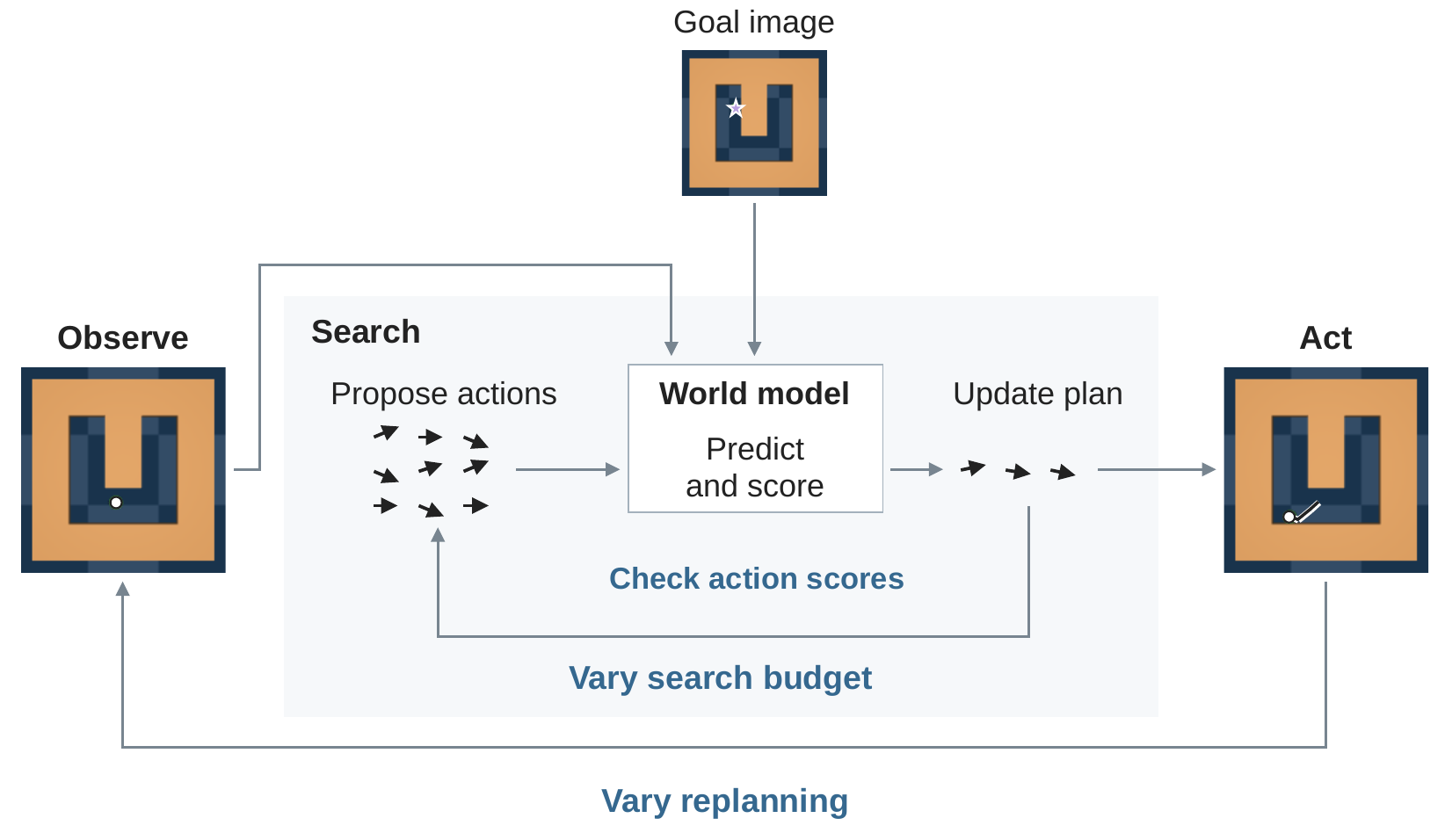}
    \caption{\textbf{Search builds a plan; replanning updates it from new observations.} The world model scores candidate action sequences by how close their predicted outcomes are to the goal in representation space. Search repeatedly samples action sequences and uses the average of the best-scored ones to update the plan. The agent then executes the first part of the plan, observes the result, and replans. Blue labels mark where we compare model scores with actual outcomes and test how search and replanning affect success. For failed plans, we also test whether any sequence from the last search round reaches the goal.}
    \label{fig:planning-loop}
\end{figure}

\paragraph{Planning with latent world models.}
\label{sec:planning}

An encoder maps an observation to a latent vector, which a predictor then rolls forward under a candidate action sequence. A planner then scores each candidate by the distance between the candidate's predicted latent and the goal observation's latent, before executing the first few actions of the resulting plan and replanning from the new observation (Figure~\ref{fig:planning-loop}). DINO-WM, JEPA-WM, V-JEPA 2-AC and related world models use such planners, usually searching over candidates with the cross-entropy method (CEM) \citep{zhou2024dinowm,terver2025jepawms,assran2025vjepa2}. CEM repeatedly samples action sequences from a Gaussian, keeps the lowest-scoring ones, called the elites, and refits the Gaussian to them. After the last round the planner executes the mean of the elites, which is itself never scored. Each candidate has a true cost, measured from the outcome it reaches rather than the one the model predicts. As a scorer, the latent distance should rank candidates the way their true cost does, and we measure the gap between the two rankings.

\paragraph{The released models.}
DINO-WM \citep{zhou2024dinowm} introduced this planner with a frozen DINOv2 image encoder and a learned predictor. JEPA-WM \citep{terver2025jepawms} studied its design choices and released checkpoints, simulated environments and evaluation configurations for four tasks, together with DINO-WM checkpoints trained as baselines. The two models share the encoder and differ in the predictor and in how it is conditioned on actions. We use these checkpoints and configurations as released, and they set the scope of our study.

\paragraph{Measurements of scorer quality.}

The closest work to ours is \citet{li2026trm}, who measure the ranking itself and find that, under a fixed encoder, the latent distance to the goal at a predicted rollout's end misranks candidates, whereas a learned reachability cost ranks the same candidates correctly. The study reports the finding for two reward-free world models, LeWorldModel (LeWM) and PLDM, on TwoRoom and TopoNav. The study also compares scorers on one shared candidate set, with the checkpoint, candidate sampler and search budget fixed, and finds on Push-T that the reachability cost improves candidate ranking more clearly than closed-loop success. We do not repeat that observation. We ask where the error sits inside a released planner and what it costs: our candidates include the ones the planner's own search produces, our true cost is the evaluation's success test, and we relate the ranking to closed-loop success. More broadly, scorers are usually judged in one of two ways. First, one line of work scores predicted futures with learned rewards or values, from PlaNet to TD-MPC2 \citep{hafner2019planet,hafner2020dreamer,hafner2023dreamerv3,hansen2022tdmpc,hansen2024tdmpc2}. There, scorer quality means how well the predicted reward or value tracks the return actually obtained. The planners we study use a latent distance and no learned value, and we evaluate that distance against executed outcomes. Second, prior work has measured how much a better model or more search improves control. For example, one-step log-likelihood of model predictions correlates only inconsistently with episode return under sampling-based planning on cartpole and half-cheetah \citep{lambert2020objective}. Similarly, evaluation-time search adds only a small boost to MuZero in most environments \citep{hamrick2021planning}.

\paragraph{Visual encoders.}

Part of the reward-free line of work takes its encoder from video rather than image pretraining \citep{bardes2024vjepa,assran2025vjepa2}. For policy learning, which encoder works best depends on the learning algorithm \citep{hu2023pretrained}, and training from scratch matched frozen pretrained features \citep{hansen2023scratch}. For latent planning, swapping the encoder inside DINO-WM changed closed-loop success, and patch features beat a single global feature vector \citep{zhou2024dinowm}. Beyond the encoder, end-to-end variants such as LeWM \citep{maes2026leworldmodel} and Delta-JEPA \citep{zhang2026deltajepa} change how the latent space is trained. However, in every study above that swaps or retrains the encoder for control, the evidence is closed-loop success or policy performance, not the ranking itself.

\paragraph{Replanning.}

Adaptive replanning changes only how often the planner replans, both for latent planners \citep{cheng2026adarep} and for map-based navigation \citep{honda2024replan}. Three other strategies change the model or the search itself. Temporal straightening retrains the encoder so that latent distance tracks geodesic distance \citep{wang2026temporal}, amortised inverse dynamics replaces the search with a learned policy \citep{nguyen2026latentgeometry}, and test-time adaptation updates the world model while the planner runs \citep{wang2026adajepa}. The five studies above report closed-loop success or task performance as evidence, and so leave the ranking unmeasured. Closed-loop success cannot stand in for the ranking, because a planner that replans at every step can correct a wrong choice before it costs an episode \citep{cheng2026adarep} and so can succeed often even when its ranking is badly wrong. We call this disagreement between the ranking and closed-loop success the ranking--success mismatch. Feedback is a legitimate part of model-predictive control, and we do not treat a planner without it as the fair setting. We use planning once as a diagnostic that shows what one choice of the score is worth before feedback corrects it. We measure the ranking directly on executed candidates and, separately, run the planner while varying how often it replans and how widely it searches.

\section{Experimental Setup}
\label{sec:setup}

Our goal is to measure the score inside the released planner, so we change as little as possible. The checkpoints, the planner and the environments are those of the release. We vary only the search budget and how often the planner replans, and we add one thing: we execute candidate action sequences and record what happens.

\paragraph{Tasks and models.}
We use the four simulated tasks of the JEPA-WM release: PointMaze \citep{fu2021d4rldatasetsdeepdatadriven} and Wall \citep{zhou2024dinowm}, which are navigation tasks, and Push-T \citep{chi2024diffusionpolicyvisuomotorpolicy} and MetaWorld reach-wall \citep{yu2021metaworldbenchmarkevaluationmultitask}, which are manipulation tasks (Figure~\ref{fig:tasks} in Appendix~\ref{app:ranking-test}). For PointMaze, Wall and Push-T we evaluate the released JEPA-WM and DINO-WM checkpoints. MetaWorld has the JEPA-WM checkpoint only, because the released DINO-WM one is trained on a different task. The two models share the DINOv2 encoder and differ in the predictor. We never train or change a weight.

\paragraph{Candidates.}
We work on the evaluation episodes themselves, 144 per task (48 under each of three seeds). At the planner's first decision in episode $i$, with observation $o_i$ and goal image $g_i$, we collect four sets $\mathcal{C}_i$ of 64 candidate action sequences. Two are the planner's own: the samples of its last CEM round, from which it forms the plan it executes, and the samples of its first round, drawn before any selection. Two do not depend on the model: action windows recorded in the released dataset from states near the episode's start (Data), and simple hand-built sequences such as constant, random and sinusoidal actions (Wide). Every candidate spans the planning horizon of 30 environment steps. The score of a candidate $c$ is the model's released planning objective,
\begin{equation}
  S_i(c) = \ell\bigl(\hat{z}_i(c),\, E(g_i)\bigr) , \label{eq:latent-distance}
\end{equation}
where $E$ is the encoder, $\hat{z}_i(c)$ is the latent that the predictor reaches from $E(o_i)$ at the end of $c$, and $\ell$ is a squared distance between latents. A lower score is better.

\begin{figure}[!t]
    \centering
    \includegraphics[width=\linewidth]{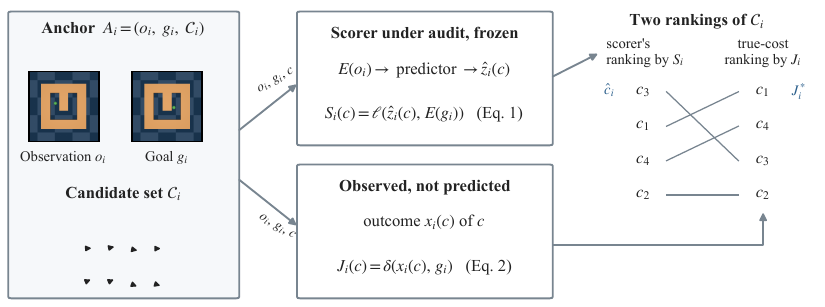}    \caption{\textbf{How we compare the score with the outcome for one episode's candidates (an \emph{anchor} in the diagram).} The scorer never receives an observed outcome $x_i(c)$ or a true cost $J_i(c)$. We compare the candidate that the scorer ranks first, $\hat{c}_i$, with the true costs of all candidates. The two frames are a PointMaze observation and goal, and the two rankings of four candidates are an illustration, not data.}
    \label{fig:audit}
\end{figure}

\paragraph{True cost.}
We execute every candidate from the episode's start state, in the environment that the evaluation itself uses (Figure~\ref{fig:audit}). To judge the outcome we need a cost that agrees with what the evaluation counts as success. Each task's success test compares a distance with a threshold, so we use their ratio,
\begin{equation}
  J_i(c) = \delta\bigl(x_i(c),\, g_i\bigr) , \label{eq:true-cost}
\end{equation}
where $x_i(c)$ is the state that $c$ reaches and $\delta$ is the distance of the task's success test divided by its threshold. A candidate succeeds exactly when $J_i(c) < 1$, and a smaller $J_i(c)$ means it ends closer to success. The score never sees $x_i(c)$ or $J_i(c)$.

\paragraph{Metrics.}
The search is steered by the lowest-scoring candidates, so we ask how good the top-scored one is. Let $\hat{c}_i$ be the top-scored candidate, $J^{*}_i$ the lowest true cost in $\mathcal{C}_i$, and $\bar{J}_i$ the mean true cost. The regret of the top-scored candidate is $J_i(\hat{c}_i) - J^{*}_i$, and a candidate picked at random would incur $\bar{J}_i - J^{*}_i$ on average. We report their ratio over all episodes, the normalised regret
\begin{equation}
  \mathrm{NR} = \frac{\sum_{i} \bigl( J_i(\hat{c}_i) - J^{*}_i \bigr)}{\sum_{i} \bigl( \bar{J}_i - J^{*}_i \bigr)} , \label{eq:nr}
\end{equation}
which is 1 for a random pick and 0 for always picking the best. We also count successes: how often the top-scored candidate succeeds, next to \emph{coverage}, the share of episodes in which some candidate succeeds.

\paragraph{Controls.}
Two controls help to read the score. The \emph{observed score} applies the same distance to the encoding of the observation that a candidate actually reaches, which bypasses the predictor. The \emph{swapped goal} scores that observation against the goal of the previous episode, which tests whether the ranking depends on the goal.

\paragraph{Closed-loop runs.}
The released checkpoints are reported to plan successfully, so we also run the planner on PointMaze, Push-T and Wall to see what drives their success. We vary two things. The planner either replans after every planner action, or plans once and executes that plan to the end of the episode. The search budget is either small, 5 CEM rounds of 64 samples, or that of the released evaluation, 30 rounds of 300 samples. In the plan-once episodes we also execute every member of the planner's final round of candidates. This tells us whether any member succeeds, whether the top-scored member succeeds, and how often a random member succeeds, next to the plan that the planner actually executes, the mean of its elites. The candidate sets above come from the small-budget plan-once episodes, and MetaWorld enters the ranking measurement only. Appendix~\ref{app:ranking-test} gives the remaining settings, Appendix~\ref{app:lewm-delta} gives the training of LeWM and Delta-JEPA, and Appendix~\ref{app:planning} the success conditions with the remaining planning settings.

\section{How Well Does Latent Distance Rank Candidates?}
\label{sec:ranking}

Closed-loop success measures the world model and the planner together, so it cannot say how well the model itself ranks. We therefore execute the candidates of Section~\ref{sec:setup} and compare the score's choice with their true costs.

\subsection{Released models}
\label{sec:ranking-released}

\begin{table}[t]
\centering
\small
\setlength{\tabcolsep}{7pt}
\begin{tabular}{llcccc}
\toprule
Task & Model & CEM last & CEM first & Data & Wide \\
\midrule
PointMaze & JEPA-WM & 0.61 & 0.39 & 0.31 & 0.31 \\
PointMaze & DINO-WM & 0.59 & 0.35 & 0.35 & 0.27 \\
Push-T & JEPA-WM & 0.58 & 0.31 & 0.27 & 0.32 \\
Push-T & DINO-WM & 0.57 & 0.34 & 0.29 & 0.47 \\
Wall & JEPA-WM & 0.38 & 0.37 & 0.05 & 0.21 \\
Wall & DINO-WM & 0.50 & 0.49 & 0.09 & 0.18 \\
\midrule
MetaWorld & JEPA-WM & 0.94 & 0.34 & 1.99 & 0.25 \\
\bottomrule
\end{tabular}
\caption{\textbf{Normalised regret of the planning score} (1 is a random pick, 0 is the best candidate). The score beats a random pick on three of four tasks, and on those tasks it ranks the planner's own final candidates least well. Each cell uses 64 executed candidates in each of 144 episodes (62 for Wall's Wide set).}
\label{tab:released}
\end{table}

Our first observation is that the score is informative: it picks better candidates than chance, although not on every task and not equally well for every set of candidates.

\paragraph{Candidate sets.}
Table~\ref{tab:released} gives the normalised regret of the planning score for each candidate set. We have two observations from it. First, the score beats a random pick on three tasks. On PointMaze, Push-T and Wall, NR is below 1 for all four candidate sets and both models, and no 95\% interval reaches 1 (Appendix~\ref{app:full}). The two models behave alike, and Push-T is ranked about as well as PointMaze.

Second, on these tasks the planner's own final candidates are ranked least well. On PointMaze and Push-T, NR is about 0.6 on the last CEM round against 0.3 to 0.4 on the first. The last round has been selected by the same score over several rounds, so its members are alike. We argue that what still separates them is what the score resolves least well.

\begin{table}[t]
\centering
\small
\setlength{\tabcolsep}{7pt}
\begin{tabular}{llccc}
\toprule
Task & Model & Predicted & Observed & Observed, swapped goal \\
\midrule
PointMaze & JEPA-WM & 0.61 & 0.06 & 1.25 \\
PointMaze & DINO-WM & 0.59 & 0.08 & 1.33 \\
Push-T & JEPA-WM & 0.58 & 0.53 & 1.05 \\
Push-T & DINO-WM & 0.57 & 0.51 & 1.12 \\
Wall & JEPA-WM & 0.38 & 0.25 & 1.15 \\
Wall & DINO-WM & 0.50 & 0.42 & 1.01 \\
\midrule
MetaWorld & JEPA-WM & 0.94 & 0.83 & 0.91 \\
\bottomrule
\end{tabular}
\caption{\textbf{Two controls on the planner's last round} (normalised regret). The ranking depends on the goal, and bypassing the predictor helps most on PointMaze. The candidates of the last CEM round are ranked by the planning score (\emph{Predicted}), by the same distance on the observation each candidate reaches (\emph{Observed}), and by that distance to the previous episode's goal.}
\label{tab:controls}
\end{table}

\paragraph{Goal and predictor.}
Table~\ref{tab:controls} studies the planner's last round with two controls. The ranking depends on the goal. Scoring the reached observations against the previous episode's goal raises NR to between 1.01 and 1.33 on the three tasks, the level of a random pick or worse. On the other candidate sets the swapped goal still gives NR between 0.38 and 0.87 (Appendix~\ref{app:full}), so part of the ranking there does not depend on the particular goal. How much the predictor limits the ranking depends on the task. Bypassing the predictor with the observed score removes most of the regret on PointMaze (0.61 to 0.06 for JEPA-WM), part of it on Wall (0.38 to 0.25), and little on Push-T (0.58 to 0.53). On Push-T, an exact prediction would not make the latent distance choose much better among the final candidates.

\paragraph{Breakdown: MetaWorld.}
The pattern does not hold on MetaWorld. There the score ranks the first CEM round and the Wide set well (0.34 and 0.25). But it does not separate the planner's final candidates from a random pick (0.94, interval 0.87 to 1.02), and on recorded windows its top pick is worse than random (1.99). None of the planner's candidates succeeds within the horizon on this task, and the recorded windows that do succeed, in 26\% of the episodes, are never the top pick. This row has one model and uses the small search budget (Appendix~\ref{app:ranking-test}).

\paragraph{Top-candidate success.}
Counting successes gives the same picture. Among the episodes whose last round contains a successful candidate, the top-scored candidate succeeds far more often than a random one: 62\% against 38\% on PointMaze, 68\% against 7\% on Push-T and 83\% against 11\% on Wall for JEPA-WM. But it does not always succeed. On PointMaze a successful candidate is present in 130 of the 144 episodes, and the score picks one in 81. Section~\ref{sec:replanning} measures what this costs in the closed loop.

\paragraph{Other encoders and end-to-end models.}
\label{sec:ranking-encoders}
We also compared image- and video-pretrained encoders, and two models trained end to end, on scripted candidates. Each model was trained once, so we report these results in Appendix~\ref{app:ranking} as observations about those checkpoints.

\section{Where Planning Success Comes From}
\label{sec:replanning}

The score often passes over a success that is available, and the released checkpoints are nonetheless reported to succeed. This section measures what carries their success. Our second observation is that a wider search and replanning both raise success with the same score, and that after a wide search the score's final choice makes little difference.

\begin{table}[t]
\centering
\small
\setlength{\tabcolsep}{5pt}
\begin{tabular}{lllccccc}
\toprule
 & & & \multicolumn{2}{c}{Closed loop} & \multicolumn{3}{c}{Final candidates of one plan} \\
\cmidrule(lr){4-5}\cmidrule(lr){6-8}
Task & Model & Budget & Replan & One plan & Coverage & Top-scored & Random \\
\midrule
PointMaze & JEPA-WM & small & 89.6 & 61.8 & 90.3 & 56.2 & 34.0 \\
PointMaze & JEPA-WM & released & 91.0 & 77.1 & 91.7 & 75.7 & 76.9 \\
PointMaze & DINO-WM & small & 82.6 & 56.9 & 88.9 & 56.9 & 33.9 \\
PointMaze & DINO-WM & released & 85.4 & 75.0 & 86.8 & 72.9 & 73.6 \\
\midrule
Push-T & JEPA-WM & small & 36.1 & 12.5 & 26.4 & 18.1 & 1.8 \\
Push-T & JEPA-WM & released & 68.1 & 67.4 & 68.8 & 66.7 & 66.7 \\
Push-T & DINO-WM & small & 41.0 & 16.7 & 23.6 & 16.7 & 2.0 \\
Push-T & DINO-WM & released & 68.8 & 61.8 & 63.9 & 61.8 & 60.4 \\
\midrule
Wall & JEPA-WM & small & 54.9 & 27.1 & 36.1 & 29.9 & 3.9 \\
Wall & JEPA-WM & released & 87.5 & 73.6 & 84.0 & 74.3 & 71.7 \\
Wall & DINO-WM & small & 31.9 & 16.0 & 25.7 & 22.9 & 2.8 \\
Wall & DINO-WM & released & 72.9 & 64.6 & 72.2 & 64.6 & 62.6 \\
\bottomrule
\end{tabular}
\caption{\textbf{Closed-loop success and the final candidates of one plan} (\%, 144 episodes). With a small search the top-scored member is far better than a random one but misses available successes. With the released search the two succeed at similar rates. \emph{One plan} executes the mean of the elites without replanning. The last three columns execute every member of that plan's final CEM round: the share of episodes in which some member succeeds (\emph{Coverage}), in which the top-scored member succeeds, and the success rate of a random member. Small is 5 rounds of 64 samples and released is 30 rounds of 300.}
\label{tab:selection}
\end{table}

\subsection{Less replanning lowers success}
\label{sec:less-replanning}

\paragraph{Replanning.}
Table~\ref{tab:selection} reports closed-loop success at the two search budgets, with and without replanning. Planning once rather than replanning lowers success in every row. At the small budget it falls from 89.6\% to 61.8\% on PointMaze, from 36.1\% to 12.5\% on Push-T and from 54.9\% to 27.1\% on Wall for JEPA-WM, and DINO-WM behaves alike. At the released budget the reduction for JEPA-WM is 13.9 points on PointMaze and on Wall, and 0.7 points on Push-T.

\paragraph{Search budget.}
A wider search helps as well, with the score unchanged. On Push-T a single plan succeeds in 12.5\% of episodes at the small budget and in 67.4\% at the released one. Search and replanning both cost computation, and the table lets us compare them: replanning at the small budget scores 1{,}920 action sequences per episode, and a single plan at the released budget scores 9{,}000. On Push-T the single large search does better (67.4\% against 36.1\%), whereas on PointMaze replanning does better with less (89.6\% against 77.1\%). A separate run that matches the total search exactly agrees for Push-T: planning once with the same total search as replanning succeeds in 70 of the 144 episodes against 52 with replanning (p = 0.0039, Appendix~\ref{app:replanning}). In sum, closed-loop success depends on how often the planner replans and on how widely it searches, by amounts that differ across tasks.

\subsection{What limits a single plan: coverage, selection or execution}
\label{sec:selection}

\begin{figure}[t]
    \centering
    \includegraphics[width=\linewidth]{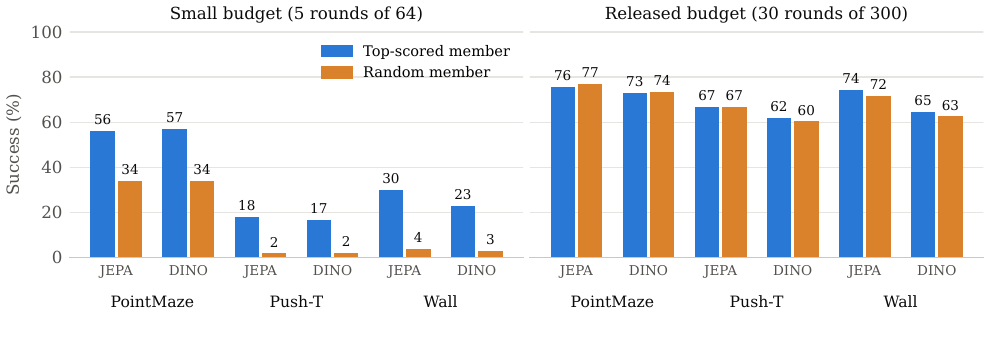}
    \caption{\textbf{Top-scored vs.\ random member} of the final candidates of one plan (success in \%, 144 episodes per bar; JEPA and DINO are JEPA-WM and DINO-WM). With a small search the top-scored member is far better. With the released search the two succeed at similar rates.}
    \label{fig:success}
\end{figure}

A single plan may fail for three reasons. The planner's final candidates may contain no successful sequence (coverage). They may contain one that the score does not rank first (selection). Or the plan that is executed, the mean of the elites, may fail where the top-scored candidate would have succeeded (execution). To separate the three, we execute every member of the final CEM round of each plan-once episode in a separate simulator instance (Table~\ref{tab:selection}, last three columns). This is a diagnostic and not a rule a planner could use, because it needs the terminal state each candidate reaches, which a planner does not have when it plans.

\paragraph{Small search.}
With a small search, the score selects well but not well enough. The top-scored member succeeds far more often than a random member in every row, by 15 to 26 points (Figure~\ref{fig:success}, left). What limits a single plan then differs by task. On PointMaze it is selection: 90.3\% of episodes have a successful member, and the top-scored member succeeds in 56.2\%. On Push-T and Wall it is coverage: only 24\% to 36\% of episodes have a successful member, and the score finds most of them.

\paragraph{Released search.}
With the released search, the top-scored member, the elite mean and a random member succeed at similar rates, within three points of one another in every row (Figure~\ref{fig:success}, right). The final candidates are not all alike: in 14 to 86 of the 144 episodes some succeed and others fail, and within those episodes the top-scored member succeeds between 2 points less and 6 points more often than a random one (Appendix~\ref{app:selection}). On PointMaze, coverage stays 14 to 16 points above the top-scored member, so successful candidates remain that the score does not pick. Executing the elite mean in place of the top-scored member changes success by up to 7 points, in either direction.

The closed-loop success rate shows none of this. PointMaze with JEPA-WM succeeds in about 90\% of episodes with replanning at both budgets, while the score misses more than a third of the available successes at one budget and does about as well as a random member at the other.

\section{Discussion}
\label{sec:discussion}

We summarize the main observations from our experiments as follows:
\begin{itemize}
    \item The score picks better candidates than chance on three of four tasks, for candidates from the planner's own search and from elsewhere.
    \item The score often passes over a successful candidate that the search has already found.
    \item The planner's own final candidates are ranked least well.
    \item A wider search and replanning both raise success with the same score.
    \item After the released search, the top-scored and a random final candidate succeed at similar rates.
\end{itemize}
Based on these observations, we provide our answers to a few questions.

\paragraph{Do latent world models select the right actions?}
Partly. The score picks better actions than chance on three of four tasks, and it misses many successes that are available.

\paragraph{Does planning success tell us?}
No. PointMaze succeeds in about 90\% of episodes both when the score misses a third of the available successes and when its final choice does about as well as a random one.

\paragraph{Is a success rate enough to evaluate a world model that plans?}
No. %
Current studies that plan with a world model judge the model by its closed-loop success rate. Few of them report a success rate without replanning \citep{zhou2024dinowm, terver2025jepawms, wang2026temporal}. Sections~\ref{sec:ranking} and~\ref{sec:replanning} show that a success rate measured with replanning does not show whether the model picks the right action. We recommend reporting the success rate without replanning, and the ranking measured directly, alongside the success rate with replanning. Replanning is a legitimate part of the method, so the single-plan number is a diagnostic and not a replacement. A reader can then tell how much of the success comes from the model picking well and how much from replanning. Two limits bound the measurements behind this recommendation. They cover two released models on simulated tasks (Appendix~\ref{app:scope}), because executing every candidate from exactly the same state needs an environment that can be reset. And the candidates are collected at the first decision of each episode.

\paragraph{Can the score be fixed?}
Not by our two attempts.
Section~\ref{sec:selection} showed that PointMaze candidates usually hold a successful one that the score does not pick. Unlike executing every candidate, a learned scorer needs no execution and can run inside the planner, so we trained two such scorers, one on PointMaze and one on Push-T. Both rank fixed candidates better than the planning objective, and neither raises the success rate (Appendix~\ref{app:fixing}, Figure~\ref{fig:first-scorer}). Better ranking on fixed candidates need not improve planning success. Other scorers may do better, and the observed-score control suggests where to look: the predictor on PointMaze, the latent distance itself on Push-T.

\section{Conclusion}
\label{sec:conclusion}

We measured, inside two released planners, whether the latent distance that scores a candidate action sequence tracks what the sequence actually achieves. The score picks better actions than chance, yet it often passes over a successful candidate that the search has already found, and planning succeeds nonetheless with the help of a wider search and replanning. With a small search, choosing correctly among the candidates already found would raise the success rate of a single plan from 62\% to 90\% on PointMaze. After the released search, the top-scored candidate and a random one succeed at similar rates. A success rate therefore says little about how well the score selects, and we hope that this direct measurement of the score will be reported next to success rates. Our study covers two released models on four simulated tasks, and MetaWorld shows that the score can also fail to separate the planner's final candidates from a random pick.

\subsection*{AI use statement}
Large language models were used to polish the manuscript and to search for related work. The authors reviewed and revised all text and take full responsibility for the content.

\subsection*{Reproducibility statement}
The released checkpoints and the simulation environments are public. Appendix~\ref{app:ranking-test} gives the settings of every measurement, with the seeds, and Appendix~\ref{app:full} the full results; Appendix~\ref{app:stats} gives the statistical procedures. The supplementary material holds the code, the episodes and the candidate sets, and the per-candidate true costs and scores.

\bibliography{references}
\bibliographystyle{preprint_style}

\appendix
\clearpage
\section{Protocol details}
\label{app:ranking-test}
\label{app:protocol}
\begin{figure}[!h]
    \centering
    \includegraphics[width=\linewidth]{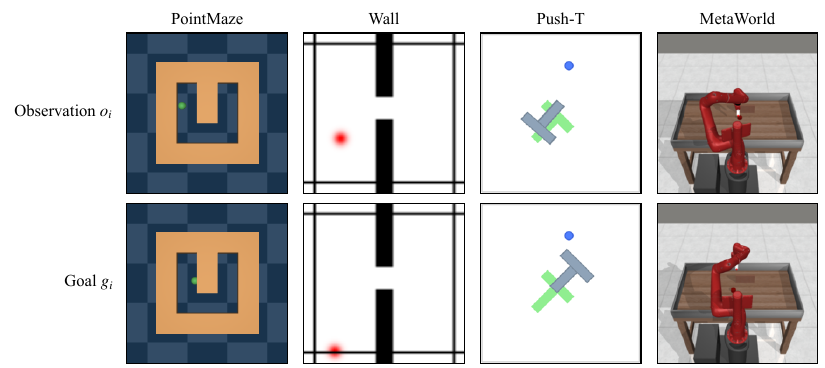}
    \caption{\textbf{One frame of an observation and of a goal from each of the four tasks.} PointMaze and Wall are navigation tasks and Push-T and MetaWorld reach-wall are manipulation tasks. Each task measures distance from the goal in a different quantity.}
    \label{fig:tasks}
\end{figure}

This appendix gives the settings behind every number of Sections~\ref{sec:ranking} and~\ref{sec:replanning}.

\subsection{Tasks and environments}
PointMaze comes from D4RL \citep{fu2021d4rldatasetsdeepdatadriven} and asks an agent to apply a force that moves a ball through a U-shaped maze to a goal position.
Wall comes from DINO-WM \citep{zhou2024dinowm} and asks an agent in one of two rooms to reach a goal in the other, through the one door in the wall between them.
Push-T comes from the Diffusion Policy work \citep{chi2024diffusionpolicyvisuomotorpolicy} and asks an agent with a round pusher to push a T-shaped block to a target position and orientation.
From MetaWorld \citep{yu2021metaworldbenchmarkevaluationmultitask} we take reach-wall, which asks a simulated Sawyer arm to move its gripper to a goal position beyond a wall.
Figure~\ref{fig:tasks} shows an observation and a goal from each task, and Appendix~\ref{app:planning} gives the observation and action spaces.
We use the environment code that ships with the JEPA-WM release for execution, rendering and the success test on every task, so each task has a single environment implementation.
MetaWorld runs on metaworld 3.0.0 with MuJoCo 3.3.0.

\subsection{Models and planning objective}
The checkpoints are the released JEPA-WM and DINO-WM checkpoints for PointMaze, Wall and Push-T and the released JEPA-WM checkpoint for MetaWorld reach-wall.
The released DINO-WM MetaWorld checkpoint is trained on the reach task and not on reach-wall, so we do not use it.
The encoder maps an observation, an image with the proprioceptive state, to a latent $z \equiv (z^{\mathrm{v}}, z^{\mathrm{p}})$, where $z^{\mathrm{v}}$ holds the DINOv2 patch features of the image and $z^{\mathrm{p}}$ is an embedding of the proprioceptive state.
The predictor advances the latent by one step for each group of five consecutive environment actions.
The function $\ell$ of Eq.~(\ref{eq:latent-distance}) is
\begin{equation}
  \ell(\hat{z}, z_g) = \frac{1}{d^{\mathrm{v}}} \bigl\lVert \hat{z}^{\mathrm{v}} - z_g^{\mathrm{v}} \bigr\rVert_2^2 + \lambda \, \frac{1}{d^{\mathrm{p}}} \bigl\lVert \hat{z}^{\mathrm{p}} - z_g^{\mathrm{p}} \bigr\rVert_2^2 ,
  \label{eq:app-score}
\end{equation}
where $d^{\mathrm{v}}$ and $d^{\mathrm{p}}$ are the numbers of entries in the two parts and $\lambda = 0.1$ is the released weight of the proprioceptive part.
Only the predicted latent at the end of the rollout enters the score, as in the released configurations.
We use these settings unchanged on every task and for both models.

\subsection{Episodes and planner settings}
Each task runs 48 episodes under each of three seeds: 70001 to 70003 on PointMaze, 71001 to 71003 on Push-T, 72001 to 72003 on Wall and 73001 to 73003 on MetaWorld.
Starts and goals depend only on the seed and the episode index, so both models and all conditions of a task see the same 144 episodes.
The planner is the released CEM planner with a horizon of six planner actions, each spanning five environment steps.
The small budget is 5 rounds of 64 samples with 8 elites, 320 scored sequences per plan, and the released budget for PointMaze, Wall and Push-T is 30 rounds of 300 samples with 10 elites, 9{,}000 per plan.
The executed plan is the mean of the Gaussian that CEM fits to the elites of its last round.
With the released momentum of zero this mean is the average of the last round's elites, and it is not itself scored.
On PointMaze, Wall and Push-T an episode lasts 30 environment steps.
\emph{Replan} plans after every planner action, six times per episode, and \emph{plan once} plans at the start and executes all six planner actions.
The success test is applied to the simulator state at the last step of the episode only.

On MetaWorld the released configuration uses 15 rounds of 300 samples, replans every three planner actions, and runs episodes of 100 environment steps.
We keep its replanning interval and use the small budget of 5 rounds of 64 samples, so that the two CEM sources of Table~\ref{tab:released} are the planner's own rounds of 64, as on the other tasks.
The MetaWorld row is therefore not measured at the released budget, and MetaWorld is not part of Table~\ref{tab:selection}.

\subsection{Candidate sources}
All sources are built at the planner's first decision of each small-budget episode, plan-once on PointMaze, Wall and Push-T.
Each holds 64 sequences of 30 environment steps.
\textbf{CEM last} and \textbf{CEM first} are the samples of the planner's last and first CEM rounds, as logged by the planner.
\textbf{Data} holds recorded action windows whose start state is nearest to the episode's start.
On PointMaze and Wall we take, from each released trajectory of a fixed pool, trajectories 896 to 1999 on PointMaze and 896 to 1919 on Wall, the window whose start position is nearest to the episode's start position, and keep the 64 nearest trajectories.
On MetaWorld we do the same over the 300 released reach-wall trajectories, with the hand position.
On Push-T we take windows from the validation trajectories other than the episode's own, in increasing order of the Push-T margin of Eq.~(\ref{eq:app-true-cost}) between the window's start state and the episode's start state.
\textbf{Wide} holds sequences that do not use the goal: the zero action, the largest and smallest recorded value along each action axis with the other axes at zero, and then, in turn until 64 are reached, one uniformly drawn action held for 30 steps, an action drawn uniformly at every step, and a sinusoid on every axis with a random period and phase.
The range of every axis is the range of the task's recorded actions.
Data and Wide do not read the goal.

\subsection{True cost}
Every candidate is executed from the episode's start state in a separate instance of the evaluation environment, and the margin of Eq.~(\ref{eq:true-cost}) is computed from the simulator state after its 30 steps:
\begin{equation}
  J(c) =
  \begin{cases}
    \lVert p(c) - p(g) \rVert_2 \,/\, 0.5 & \text{PointMaze,} \\
    \lVert p(c) - p(g) \rVert_2 \,/\, 4.5 & \text{Wall,} \\
    \max\bigl( \lVert q(c) - q(g) \rVert_2 / 20 ,\; \lvert \theta(c) - \theta(g) \rvert / (\pi/9) \bigr) & \text{Push-T,} \\
    \lVert h(c) - h(g) \rVert_2 \,/\, 0.05 & \text{MetaWorld.}
  \end{cases}
  \label{eq:app-true-cost}
\end{equation}
Here, $p$ is the two-dimensional position, $q$ is the four-dimensional vector of pusher and block positions, $\theta$ is the block angle with the difference wrapped to $[0, \pi]$, and $h$ is the gripper position, the midpoint of the two fingers.
In each case the thresholds are those of the released success test, so $J(c) < 1$ if and only if the test passes.
On MetaWorld the released evaluation reports the simulator's own success flag, which is true when the gripper is within 0.05 of the goal, and $J$ is the simulator's gripper-to-goal distance divided by 0.05.
For every executed candidate we checked that $J < 1$ agrees with the success test, and it agrees in every case.

\subsection{Controls}
The observed score renders the state that a candidate reaches after its 30 steps, encodes the observation with the model's encoder, and applies Eq.~(\ref{eq:app-score}) to this encoding and the goal encoding.
The swapped-goal score applies the same computation with the goal encoding of the previous episode of the same run.
The first episode of each run has no previous goal, so the swapped-goal results use 141 of the 144 episodes.

\subsection{Statistics}
NR is the ratio of sums in Eq.~(\ref{eq:nr}), with the first candidate taken on ties in the score.
Appendix~\ref{app:stats} gives the intervals.

\subsection{Checks and disclosures}
\textbf{Replay.} For every plan-once episode on PointMaze, Wall and Push-T we restore the logged start state, re-execute the logged plan, and compare with the logged outcome.
The start state and the outcome are reproduced in every episode.

\textbf{Wall.} In the released Wall environment a purely vertical move into the border wall returns an undefined position, because a slope is computed with a zero denominator.
This affects the two Wide candidates that move along the vertical axis at its largest and smallest value, in every Wall episode.
They have no outcome and are excluded, so Wall's Wide sets hold 62 of the 64 candidates.
No other candidate is affected.

\textbf{MetaWorld.} Closed-loop MetaWorld episodes are not bit-identical across repeated runs in our setup.
All candidates of an episode are scored and executed within one run from the same restored state, and we verify within each run that re-executing the first executed block lands on the logged state.

\section{Planning details}
\label{app:planning}

In the released configurations every environment is observed as an RGB image of 224 by 224 pixels together with the simulator's proprioceptive state, and Wall draws its image at 65 by 65 pixels before resizing it. On PointMaze the action is a two-dimensional force on the ball, with each component in the range $[-1, 1]$. On Wall the action is a two-dimensional displacement of the agent, each component in the same range and scaled by two, and a move that meets the wall stops there unless the move passes through the door. The Wall layout is fixed, with one wall position and one door position, and the start and goal of an episode lie on opposite sides of the wall. In the copy of the Push-T environment that JEPA-WM ships, the action is a two-dimensional target position for the pusher, relative to its current position. The scripted-candidate audit of Appendix~\ref{app:ranking} executes its Push-T candidates in the gym-pusht package's implementation of the same environment, whereas the runs of Appendix~\ref{app:ranking-test} use JEPA-WM's copy. On MetaWorld the task is the reach-wall variant on a simulated Sawyer arm, and the task's state vector includes the goal position. The action has four components in $[-1, 1]$: a three-dimensional displacement of the gripper and a gripper command. MetaWorld itself counts reach-wall as solved when the centre of the gripper lies within 0.05 of the goal position, and we use that condition, which the simulator reports as its own success flag. A MetaWorld episode has 100 steps.

The closed-loop runs use CEM with a planning horizon of six planner actions (each spanning five environment steps), at the small budget of five rounds of 64 samples with eight elites and at the released budget of 30 rounds of 300 samples with ten elites, the top-scored samples kept at each round. The plan executed after each search is the mean of the Gaussian that CEM fits to the elites of its last round, not one of the sampled candidates. CEM therefore scores 320 action sequences per plan at the small budget and 9{,}000 at the released one. Each task runs 48 episodes under each of three seeds, 70001 to 70003 on PointMaze, 71001 to 71003 on Push-T, 72001 to 72003 on Wall and 73001 to 73003 on MetaWorld, giving 144 episodes per task under each condition, whereas the released evaluation runs 96 episodes under seed 1.

A closed-loop episode runs 30 environment steps, which the planner covers in six planner actions, and the two replanning intervals are one planner action and six planner actions. On PointMaze the start and goal of an episode are drawn with a per-episode seed that depends only on one of the three seeds above and on the episode index. On Push-T a random-number generator that advances only with the episode index picks a segment of the released Push-T dataset, and the start and goal come from that segment, so the same index gives the same episode under both intervals. On PointMaze an episode succeeds when its final position lies within 0.5 of the goal position. On Push-T an episode succeeds when the four-dimensional vector of agent and T-shaped block positions lies within 20 of the goal's and the block angle lies within $\pi/9$ of the goal angle. Both success conditions are applied to the simulator state at the last step of the episode only, so an episode that reaches the goal and then leaves the goal counts as a failure.

\paragraph{Settings and uncertainty for Figure~\ref{fig:teaser}.}
Panel (a) compares two candidates from the last CEM round of the same search on PointMaze. The top-scored candidate has a latent distance of 0.614 and misses the goal, whereas the candidate that reaches the goal has a latent distance of 0.627, with lower distances preferred. In panels (b) and (c), the smaller budget is our reduced search budget, with 64 samples per round, five rounds and eight elites, and the larger budget is that of the released evaluation, with 300 samples per round, 30 rounds and ten elites. Panel (b) compares these budgets on Push-T with one plan executed over 30 environment steps. Panel (c) uses the larger budget on PointMaze and compares one plan with replanning every five environment steps. Each comparison uses 144 paired episodes across three seeds. Push-T success is 12.5\% under the smaller budget and 67.4\% under the larger budget, with episode-level 95\% Wilson intervals of [8.1, 18.9] and [59.3, 74.5] percent. PointMaze success is 77.1\% with one plan and 91.0\% with replanning, with corresponding intervals of [69.6, 83.2] and [85.2, 94.6] percent.

\paragraph{Illustration in Figure~\ref{fig:planning-loop}.}
The PointMaze scenes come from a simulator replay. The scenes show the initial observation and the state after five environment steps; the goal observation shown is rendered from the episode's recorded goal state. The action symbols are schematic. Within the search loop, CEM updates its sampling distribution using the mean and standard deviation of the top-scored action sequences.

\section{Full results}
\label{app:full}
\label{app:selection}

Table~\ref{tab:nr-full} extends Table~\ref{tab:released} with intervals, coverage and top-candidate success.
Table~\ref{tab:controls-full} gives the two controls for every candidate source.
Table~\ref{tab:success-full} gives intervals for Table~\ref{tab:selection} and the episodes in which the members of a plan's final CEM round disagree.

The top-scored member of that round succeeds more often than a random member at the small budget and about as often at the released one.
At the small budget the difference is between $+14.7$ and $+26.0$ points in all six rows of Table~\ref{tab:selection}, and the paired 95\% bootstrap interval over episodes excludes zero in every row.
At the released budget the difference is between $-1.2$ and $+2.6$ points, and the interval includes zero in five of the six rows.
The exception is Push-T with DINO-WM, at $+1.4$ points with an interval of $[+0.4, +2.5]$. The share of episodes in which the top-scored candidate of the last CEM round is exactly the best one is 4.9\% to 30.6\% on PointMaze, Push-T and Wall and 3.5\% on MetaWorld, against 1.6\% to 1.7\% for a random pick.

\begin{table}[!ht]
\centering
\footnotesize
\setlength{\tabcolsep}{4pt}
\begin{tabular}{lllcccccc}
\toprule
 & & & & & & \multicolumn{3}{c}{Covered episodes} \\
\cmidrule(lr){7-9}
Task & Model & Source & NR [95\% interval] & Coverage & Top succ. & $n$ & Top & Random \\
\midrule
PointMaze & JEPA-WM & CEM last & 0.61 [0.52, 0.70] & 90.3 & 56.2 & 130 & 62.3 & 37.7 \\
PointMaze & JEPA-WM & CEM first & 0.39 [0.32, 0.46] & 79.9 & 44.4 & 115 & 55.7 & 12.1 \\
PointMaze & JEPA-WM & Data & 0.31 [0.25, 0.37] & 81.2 & 46.5 & 117 & 57.3 & 11.5 \\
PointMaze & JEPA-WM & Wide & 0.31 [0.26, 0.38] & 88.2 & 49.3 & 127 & 55.9 & 8.4 \\
\midrule
PointMaze & DINO-WM & CEM last & 0.59 [0.51, 0.68] & 88.9 & 56.9 & 128 & 64.1 & 38.1 \\
PointMaze & DINO-WM & CEM first & 0.35 [0.29, 0.42] & 79.9 & 51.4 & 115 & 64.3 & 12.1 \\
PointMaze & DINO-WM & Data & 0.35 [0.28, 0.42] & 81.2 & 50.0 & 117 & 61.5 & 11.5 \\
PointMaze & DINO-WM & Wide & 0.27 [0.22, 0.33] & 88.2 & 50.7 & 127 & 57.5 & 8.4 \\
\midrule
Push-T & JEPA-WM & CEM last & 0.58 [0.50, 0.66] & 26.4 & 18.1 & 38 & 68.4 & 6.8 \\
Push-T & JEPA-WM & CEM first & 0.31 [0.25, 0.38] & 4.2 & 2.8 & 6 & 66.7 & 2.6 \\
Push-T & JEPA-WM & Data & 0.27 [0.21, 0.33] & 11.8 & 6.9 & 17 & 58.8 & 1.7 \\
Push-T & JEPA-WM & Wide & 0.32 [0.27, 0.38] & 1.4 & 1.4 & 2 & 100.0 & 1.6 \\
\midrule
Push-T & DINO-WM & CEM last & 0.57 [0.49, 0.64] & 23.6 & 16.7 & 34 & 70.6 & 8.5 \\
Push-T & DINO-WM & CEM first & 0.34 [0.28, 0.41] & 4.2 & 3.5 & 6 & 83.3 & 2.6 \\
Push-T & DINO-WM & Data & 0.29 [0.22, 0.36] & 11.8 & 5.6 & 17 & 47.1 & 1.7 \\
Push-T & DINO-WM & Wide & 0.47 [0.38, 0.56] & 1.4 & 1.4 & 2 & 100.0 & 1.6 \\
\midrule
Wall & JEPA-WM & CEM last & 0.38 [0.30, 0.46] & 36.1 & 29.9 & 52 & 82.7 & 10.7 \\
Wall & JEPA-WM & CEM first & 0.37 [0.30, 0.44] & 1.4 & 1.4 & 2 & 100.0 & 1.6 \\
Wall & JEPA-WM & Data & 0.05 [0.03, 0.08] & 23.6 & 21.5 & 34 & 91.2 & 1.9 \\
Wall & JEPA-WM & Wide & 0.21 [0.16, 0.28] & 8.3 & 7.6 & 12 & 91.7 & 1.7 \\
\midrule
Wall & DINO-WM & CEM last & 0.50 [0.40, 0.59] & 25.7 & 22.9 & 37 & 89.2 & 11.0 \\
Wall & DINO-WM & CEM first & 0.49 [0.41, 0.57] & 1.4 & 0.7 & 2 & 50.0 & 1.6 \\
Wall & DINO-WM & Data & 0.09 [0.06, 0.12] & 23.6 & 22.9 & 34 & 97.1 & 1.9 \\
Wall & DINO-WM & Wide & 0.18 [0.13, 0.23] & 8.3 & 6.9 & 12 & 83.3 & 1.7 \\
\midrule
MetaWorld & JEPA-WM & CEM last & 0.94 [0.87, 1.02] & 0.0 & 0.0 & 0 & -- & -- \\
MetaWorld & JEPA-WM & CEM first & 0.34 [0.28, 0.39] & 0.0 & 0.0 & 0 & -- & -- \\
MetaWorld & JEPA-WM & Data & 1.99 [1.86, 2.14] & 26.4 & 0.0 & 38 & 0.0 & 7.6 \\
MetaWorld & JEPA-WM & Wide & 0.25 [0.20, 0.31] & 9.7 & 4.9 & 14 & 50.0 & 1.7 \\
\bottomrule

\end{tabular}
\caption{\textbf{Ranking by the planning score, all candidate sources.} Coverage is the share of the 144 episodes in which some candidate succeeds and \emph{Top succ.} the share in which the top-scored candidate succeeds (\%). The last three columns restrict to the $n$ covered episodes and give the success of the top-scored candidate and of a random candidate (\%).}
\label{tab:nr-full}
\end{table}

\begin{table}[!ht]
\centering
\footnotesize
\setlength{\tabcolsep}{4pt}
\begin{tabular}{lllccc}
\toprule
Task & Model & Source & Predicted & Observed & Observed, swapped goal \\
\midrule
PointMaze & JEPA-WM & CEM last & 0.61 [0.52, 0.70] & 0.06 [0.04, 0.10] & 1.25 [1.11, 1.39] \\
PointMaze & JEPA-WM & CEM first & 0.39 [0.32, 0.46] & 0.07 [0.04, 0.11] & 0.87 [0.78, 0.97] \\
PointMaze & JEPA-WM & Data & 0.31 [0.25, 0.37] & 0.07 [0.04, 0.10] & 0.87 [0.78, 0.96] \\
PointMaze & JEPA-WM & Wide & 0.31 [0.26, 0.38] & 0.04 [0.02, 0.06] & 0.78 [0.70, 0.87] \\
\midrule
PointMaze & DINO-WM & CEM last & 0.59 [0.51, 0.68] & 0.08 [0.05, 0.11] & 1.33 [1.17, 1.48] \\
PointMaze & DINO-WM & CEM first & 0.35 [0.29, 0.42] & 0.10 [0.06, 0.14] & 0.87 [0.78, 0.96] \\
PointMaze & DINO-WM & Data & 0.35 [0.28, 0.42] & 0.08 [0.05, 0.11] & 0.84 [0.76, 0.93] \\
PointMaze & DINO-WM & Wide & 0.27 [0.22, 0.33] & 0.04 [0.02, 0.06] & 0.78 [0.70, 0.87] \\
\midrule
Push-T & JEPA-WM & CEM last & 0.58 [0.50, 0.66] & 0.53 [0.45, 0.61] & 1.05 [0.94, 1.17] \\
Push-T & JEPA-WM & CEM first & 0.31 [0.25, 0.38] & 0.31 [0.25, 0.38] & 0.61 [0.52, 0.70] \\
Push-T & JEPA-WM & Data & 0.27 [0.21, 0.33] & 0.33 [0.26, 0.41] & 0.74 [0.66, 0.83] \\
Push-T & JEPA-WM & Wide & 0.32 [0.27, 0.38] & 0.36 [0.31, 0.43] & 0.38 [0.33, 0.44] \\
\midrule
Push-T & DINO-WM & CEM last & 0.57 [0.49, 0.64] & 0.51 [0.44, 0.58] & 1.12 [1.02, 1.24] \\
Push-T & DINO-WM & CEM first & 0.34 [0.28, 0.41] & 0.35 [0.27, 0.43] & 0.61 [0.52, 0.72] \\
Push-T & DINO-WM & Data & 0.29 [0.22, 0.36] & 0.30 [0.23, 0.37] & 0.74 [0.65, 0.83] \\
Push-T & DINO-WM & Wide & 0.47 [0.38, 0.56] & 0.40 [0.34, 0.46] & 0.38 [0.33, 0.44] \\
\midrule
Wall & JEPA-WM & CEM last & 0.38 [0.30, 0.46] & 0.25 [0.18, 0.34] & 1.15 [1.04, 1.27] \\
Wall & JEPA-WM & CEM first & 0.37 [0.30, 0.44] & 0.44 [0.36, 0.51] & 0.78 [0.70, 0.87] \\
Wall & JEPA-WM & Data & 0.05 [0.03, 0.08] & 0.04 [0.02, 0.07] & 0.69 [0.61, 0.76] \\
Wall & JEPA-WM & Wide & 0.21 [0.16, 0.28] & 0.16 [0.11, 0.21] & 0.71 [0.63, 0.80] \\
\midrule
Wall & DINO-WM & CEM last & 0.50 [0.40, 0.59] & 0.42 [0.33, 0.52] & 1.01 [0.91, 1.11] \\
Wall & DINO-WM & CEM first & 0.49 [0.41, 0.57] & 0.48 [0.41, 0.56] & 0.77 [0.68, 0.85] \\
Wall & DINO-WM & Data & 0.09 [0.06, 0.12] & 0.04 [0.02, 0.07] & 0.69 [0.62, 0.77] \\
Wall & DINO-WM & Wide & 0.18 [0.13, 0.23] & 0.17 [0.12, 0.23] & 0.75 [0.66, 0.83] \\
\midrule
MetaWorld & JEPA-WM & CEM last & 0.94 [0.87, 1.02] & 0.83 [0.76, 0.89] & 0.91 [0.83, 0.98] \\
MetaWorld & JEPA-WM & CEM first & 0.34 [0.28, 0.39] & 0.33 [0.28, 0.37] & 0.41 [0.35, 0.46] \\
MetaWorld & JEPA-WM & Data & 1.99 [1.86, 2.14] & 1.75 [1.62, 1.89] & 1.96 [1.83, 2.09] \\
MetaWorld & JEPA-WM & Wide & 0.25 [0.20, 0.31] & 0.48 [0.40, 0.56] & 0.53 [0.45, 0.62] \\
\bottomrule

\end{tabular}
\caption{\textbf{Normalised regret under the two controls, all candidate sources}, with 95\% bootstrap intervals. The swapped-goal column uses 141 episodes.}
\label{tab:controls-full}
\end{table}

\begin{table}[!ht]
\centering
\footnotesize
\setlength{\tabcolsep}{4pt}
\begin{tabular}{lllccccc}
\toprule
 & & & & & \multicolumn{3}{c}{Episodes with mixed outcomes} \\
\cmidrule(lr){6-8}
Task & Model & Budget & Replan & Plan once & $n$ & Top-scored & Random \\
\midrule
PointMaze & JEPA-WM & small & 89.6 [83.5, 93.6] & 61.8 [53.7, 69.3] & 130 & 62.3 & 37.7 \\
PointMaze & JEPA-WM & released & 91.0 [85.2, 94.6] & 77.1 [69.6, 83.2] & 82 & 72.0 & 74.1 \\
PointMaze & DINO-WM & small & 82.6 [75.6, 88.0] & 56.9 [48.8, 64.7] & 128 & 64.1 & 38.1 \\
PointMaze & DINO-WM & released & 85.4 [78.7, 90.3] & 75.0 [67.3, 81.4] & 80 & 75.0 & 76.2 \\
Push-T & JEPA-WM & small & 36.1 [28.7, 44.2] & 12.5 [8.1, 18.9] & 38 & 68.4 & 6.8 \\
Push-T & JEPA-WM & released & 68.1 [60.1, 75.1] & 67.4 [59.3, 74.5] & 14 & 78.6 & 79.1 \\
Push-T & DINO-WM & small & 41.0 [33.3, 49.1] & 16.7 [11.5, 23.6] & 34 & 70.6 & 8.5 \\
Push-T & DINO-WM & released & 68.8 [60.8, 75.8] & 61.8 [53.7, 69.3] & 32 & 90.6 & 84.4 \\
Wall & JEPA-WM & small & 54.9 [46.7, 62.8] & 27.1 [20.5, 34.9] & 52 & 82.7 & 10.7 \\
Wall & JEPA-WM & released & 87.5 [81.1, 91.9] & 73.6 [65.9, 80.1] & 86 & 83.7 & 79.3 \\
Wall & DINO-WM & small & 31.9 [24.9, 39.9] & 16.0 [10.9, 22.8] & 37 & 89.2 & 11.0 \\
Wall & DINO-WM & released & 72.9 [65.1, 79.5] & 64.6 [56.5, 71.9] & 66 & 83.3 & 79.0 \\
\bottomrule

\end{tabular}
\caption{\textbf{Closed-loop success (\%) with 95\% Wilson intervals, and the plan-once episodes in which some members of the final CEM round succeed and others fail.} Within these $n$ episodes, the last two columns give the success of the top-scored member and of a random member (\%). \emph{Plan once} executes the mean of the elites.}
\label{tab:success-full}
\end{table}

\section{LeWM and Delta-JEPA}
\label{app:lewm-delta}
For the LeWM and Delta-JEPA models we trained, the encoder is a randomly initialized ViT-Tiny with $14\times14$ patches and 192-dimensional latent vectors.
Both models use a six-layer Transformer predictor with 16 attention heads, 64 dimensions per head, and a feed-forward width of 2,048.
The predictor uses causal attention and incorporates action embeddings through adaptive layer normalization.
We use $224\times224$ images and a context of three observations spaced five environment steps apart.
The five two-dimensional actions between successive observations are concatenated into a 10-dimensional action input.

Delta-JEPA reconstructs these actions with a three-layer non-causal Transformer decoder containing five learned queries, one per environment step.
The decoder has eight attention heads with 64 dimensions per head and a feed-forward width of 512.
The difference between successive latent vectors is projected to 512 dimensions and conditions every decoder block through adaptive layer normalization.

Both models minimize a prediction loss $\mathcal{L}_{\mathrm{pred}}$, the mean squared error between predicted and encoded latent vectors at the next model step.
LeWM uses $\mathcal{L}_{\mathrm{pred}}+0.09\mathcal{L}_{\mathrm{SIGReg}}$, where $\mathcal{L}_{\mathrm{SIGReg}}$ encourages an isotropic Gaussian distribution of latent vectors.
Delta-JEPA uses $\mathcal{L}_{\mathrm{pred}}+10\mathcal{L}_{\mathrm{action}}$, where $\mathcal{L}_{\mathrm{action}}$ is the mean squared error between the reconstructed and normalized recorded action inputs.
The encoder and predictor are optimized jointly in both models, together with the action decoder in Delta-JEPA.

We use LeWM's Push-T expert dataset with a random 90\%/10\% training--validation split, generated with seed 3072.
Images are normalized using ImageNet statistics.
Each action coordinate is standardized using its mean and standard deviation over the full source dataset, computed before the split.

Both models are trained for 50 epochs using AdamW, a batch size of 128, a learning rate of $5\times10^{-5}$, and weight decay of $10^{-3}$.
We use the training framework's linear-warmup and cosine-annealing scheduler, updated once per epoch.
Training uses bfloat16 precision and a gradient-clipping threshold of 1.0.
Evaluation uses the epoch-50 checkpoints.

\clearpage
\section{Scripted candidates under the dense-reward cost}
\label{app:ranking}

Appendix~\ref{app:full} gives the 95\% intervals of Table~\ref{tab:released}.
This appendix reports a separate analysis of the same released checkpoints.
It scores candidate action sequences that are scripted and fixed in advance, and on Push-T it judges them by a cost built from the environment's dense reward, one minus the highest reward during the rollout.
Both the candidates and the cost differ from those of Table~\ref{tab:released}: there the score beats a random pick on the manipulation candidates, and here it ranks them no better than a random pick.
The difference in outcome cannot be attributed to the candidates or to the cost alone.

\subsection{Candidates and metrics}
We call a start state, its goal observation and its fixed candidate set an \emph{anchor}.
The candidates come from three sources, and how many an anchor offers depends on the task: twelve on PointMaze and Wall and thirty-two in Push-T and MetaWorld. On PointMaze and Wall each candidate is the action sequence that a released trajectory follows after the time step of the last frame of the anchor's observation, and the pool of released trajectories excludes the trajectories the anchors are drawn from. An anchor's twelve candidates come from the trajectories whose positions lie nearest to the anchor's at the time step of the last frame of its observation. In Push-T the candidates are rollouts from sixteen scripted push heuristics, two per heuristic, each one executed from a simulator reset with the seed that generated the anchor. In MetaWorld the candidates come in sixteen kinds, two of each. The first is the scripted expert's own action prefix, cut to the thirty actions a candidate holds. Seven more modify that prefix, adding Gaussian noise of scale 0.05, 0.15 or 0.35, reversing the order of its actions, negating its x, y and z components, or setting its first or its second half to zero. The other eight do not use the expert. They are no action, an action drawn at random at each step or drawn once and held, the largest action along each of the $+x$, $-x$, $+y$ and $-y$ directions, and a sinusoid on the x, y and z components. The expert rollout that defines the goal state runs longer than thirty actions, so the expert's prefix usually does not reach the goal, and the prefix still holds the anchor's lowest true cost on 69 of the 750 anchors. Relative action values outside the range the Push-T checkpoint saw in training make up 2.61\% of that task's candidate values, and we leave them unclipped. On every task the true cost comes from executing the candidate from the anchor's state. The true cost is, on each task,
\begin{equation}
  J_i(c) =
  \begin{cases}
    \bigl\lVert p(c) - p(g_i) \bigr\rVert_2 & \text{on PointMaze and Wall,} \\
    1 - \max_{1 \le t \le 30} r_t(c) & \text{on Push-T,} \\
    \bigl\lVert s(c) - s(g_i) \bigr\rVert_2 & \text{on MetaWorld.}
  \end{cases}
  \label{eq:scripted-cost}
\end{equation}
Here, $p(c)$ and $p(g_i)$ are two-dimensional positions and $r_t(c)$ is the Push-T dense reward after the $t$-th of the candidate's thirty actions. Push-T's true cost therefore reads the candidate's best step, while the planning objective predicts its last. Scoring the same audit by $1 - r_{30}(c)$ instead retains 596 of the 750 anchors and gives a Hit@1 of 3.9\% against a random ranking's 5.9\%, so the planning objective falls below a random ranking under either definition. Here, $s(c)$ and $s(g_i)$ are MetaWorld simulator state vectors at the end of the candidate's execution and at the end of an expert rollout from the same start. On PointMaze and Wall, $p(c)$ is the position reached when the candidate's ten actions are executed from the anchor's state at the last frame of its observation $o_i$, a state that holds position and velocity on PointMaze, and $p(g_i)$ is the position in the anchor's own trajectory ten environment steps after that frame. The PointMaze and Wall goal observation $g_i$ is the frame of the anchor's own trajectory in which $p(g_i)$ is measured. PointMaze is executed in the released MuJoCo maze model stepped with MuJoCo 2.1.2, and Wall with the transition of the released Wall data generator. Replayed from their stored states, both simulators reproduce every stored trajectory the audit uses, to within $10^{-5}$ in position on PointMaze and exactly on Wall. The released Wall frames follow a different rule, that of the evaluation environment, which moves the agent by twice the action, and the Wall true costs here follow the recorded states. For scoring, a candidate's actions are spliced onto the anchor's own observation and actions, so the score and the true cost describe the same start state and the same actions. The Push-T dense reward is the block's overlap with the target region, as a fraction of that region, divided by $0.95$ and capped at one. The Push-T goal observation is a rendering of the scene with the block placed at its target pose. The MetaWorld goal observation is the observation at the end of the expert rollout that gives $s(g_i)$.

Three metrics report the ranking of an anchor's candidates. Hit@1 is the fraction of anchors whose top-scored candidate ties for the lowest true cost in its own anchor, and the wrong-anchor rate is one minus Hit@1. Top-1 regret is the mean of $J_i(\hat{c}_i) - J^{*}_i$ over the anchors, where $\hat{c}_i$ is the top-scored candidate and $J^{*}_i$ the lowest true cost of the anchor, so it is read in the task's own units. Spearman is the per-anchor rank correlation between score and true cost, averaged over anchors, and is zero for a random ranking. A random ranking is scored by the same tie rule, so its Hit@1 exceeds one over the candidate count: it reaches 8.3\% on PointMaze, 9.6\% on Wall, 5.8\% on Push-T and 6.0\% on MetaWorld. On Push-T and MetaWorld an anchor whose candidates all share one true cost is dropped. Both top-1 regret and Hit@1 are measured against the best candidate in the same anchor, so neither depends on how good an anchor's candidates are.

The PointMaze and Wall numbers we report come from a held-out split fixed in advance, which holds 256 anchors taken from released trajectories 640 to 895, an anchor per trajectory. The split's candidates come from released trajectories 896 to 1999 on PointMaze and 896 to 1919 on Wall. Candidates on Push-T and MetaWorld are scripted instead, so the contrast between the navigation and manipulation tasks is also a contrast between candidates drawn from data and scripted ones, which we do not separate.

The score of Table~\ref{tab:scripted} uses the proprioceptive weight $\lambda = 1$ on PointMaze and Wall, against the released $\lambda = 0.1$, an unweighted sum set before the held-out split was created and not tuned, and the released weight on Push-T and MetaWorld. With $\lambda = 0.1$ the PointMaze Hit@1 is 30.9\% rather than 31.6\%, the top-1 regret 0.155 [0.13, 0.18] rather than 0.132 against a random 0.445, and the Spearman correlation 0.500 rather than 0.581. On Wall the proprioceptive term makes up about 0.002\% of the score, so the Wall values of Table~\ref{tab:scripted} are unchanged at their printed precision.

The six checkpoints are files of the facebook/jepa-wms release: the four JEPA-WM files for PointMaze, Wall, Push-T and MetaWorld and the two DINO-WM files for PointMaze and Wall, each loaded with the released evaluation configuration for its task. Each checkpoint is loaded for inference only and without an optimiser, no gradients are computed while scoring, and no script writes to a checkpoint file. Where a later section trains a scorer or a head, the training touches that scorer or head alone and leaves the six checkpoints as released. In all six checkpoints the image encoder is the publicly available DINOv2 ViT-S/14, which the release code loads in the same way for JEPA-WM and for DINO-WM. JEPA-WM and DINO-WM therefore differ after the encoder, in the predictor and in whether the action enters that predictor as a token or as a feature.

\subsection{Results}

\begin{table}[!ht]
\centering
\footnotesize
\setlength{\tabcolsep}{4pt}
\begin{tabular}{llrrcr}
\toprule
Task & Anchors & Hit@1 & Wrong & Regret [95\% CI] & Spearman \\
\midrule
\multicolumn{6}{l}{\emph{Twelve candidates per anchor}} \\
PointMaze & 256 & 31.6\% & 0.684 & 0.132 [0.11, 0.15] & 0.581 \\
Wall & 256 & 50.8\% & 0.492 & 1.146 [0.92, 1.38] & 0.623 \\
Random (PointMaze) & --- & 8.3\% & 0.917 & 0.445 [0.42, 0.47] & 0.000 \\
Random (Wall) & --- & 9.6\% & 0.904 & 6.546 [6.25, 6.84] & 0.000 \\
\midrule
\multicolumn{6}{l}{\emph{Thirty-two candidates per anchor}} \\
Push-T & 618 / 750 & 2.43\% & 0.976 & 0.222 [0.21, 0.24] & 0.065 \\
MetaWorld & 750 / 750 & 0.0\% & 1.000 & 0.283 [0.28, 0.29] & -0.597 \\
Random (Push-T) & --- & 5.8\% & 0.942 & 0.207 [0.20, 0.22] & 0.000 \\
Random (MetaWorld) & --- & 6.0\% & 0.940 & 0.257 [0.26, 0.26] & 0.000 \\
\bottomrule
\end{tabular}
\caption{\textbf{Ranking of scripted candidates fixed in advance, by each task's released checkpoint under its own planning objective, with a random row per task.} Hit@1 and the wrong-anchor rate sum to one. The random rows give what a random ranking of the same candidates reaches. On Push-T and MetaWorld the two counts are the anchors retained and the anchors built, since an anchor whose candidates all share one true cost is dropped. The regret of a task is in that task's own units. Under these candidates and these costs the checkpoints rank no better than a random ranking on both manipulation tasks.}
\label{tab:scripted}
\end{table}

The JEPA-WM checkpoints misrank most anchors on three of the four tasks (Table~\ref{tab:scripted}). On the two manipulation tasks, Push-T and MetaWorld, the planning objective's Hit@1 falls below what a random ranking reaches. On Push-T the planning objective's Hit@1 is 2.43\%, against the 5.8\% of a random ranking. On MetaWorld the planning objective never puts a best candidate first across all 750 anchors, a Hit@1 of 0.0\% against a random ranking's 6.0\%; its top-scored candidate holds the gripper still or acts at random on 745 of the 750 anchors.

By contrast, on the two navigation tasks the planning objective beats a random ranking's Hit@1 of 8.3\% and 9.6\%, reaching 31.6\% on PointMaze and 50.8\% on Wall over 256 anchors each. Executing each candidate under the transition that the released Wall frames follow instead gives a Hit@1 of 62.5\% against a random ranking's 9.0\%. On 32 anchors per task scored in one run, DINO-WM misranks most PointMaze anchors too but only 9 of the 32 Wall anchors. Scoring the observation a Push-T candidate actually reaches, in place of the predicted one, raises Hit@1 from 2.4\% to 20.7\%, so under this cost the ranking of Push-T candidates rises above a random ranking once the predictor is bypassed.

The DINO-WM comparison uses the 32-anchor protocol under which all four PointMaze and Wall checkpoints were scored in one run, and not the 256-anchor held-out split behind Table~\ref{tab:scripted}. Each candidate of those 32 anchors is executed from its anchor's full state, as in Table~\ref{tab:scripted}. On those anchors DINO-WM reaches a Hit@1 of 21.9\% against JEPA-WM's 28.1\% on PointMaze, and 71.9\% against 40.6\% on Wall. The exact paired test on the paired outcomes returns a p-value of 0.69 on PointMaze and 0.0020 on Wall, where DINO-WM alone is right on 10 anchors and JEPA-WM alone on none. One anchor is worth 3.1 points at this size, so the two models are separated on Wall and not on PointMaze.

The planning objective mixes the geometry of the latent distance with the predictor's error. To separate the two, we also score each candidate with the same planning objective applied to the encoded observation that the candidate actually reaches, with the goal unchanged. On Push-T this observed score reaches a Hit@1 of 20.7\% against the planning objective's 2.4\% and a random ranking's 5.8\%, over the same 618 anchors. Its top-1 regret is 0.124 [0.113, 0.135], against 0.222 [0.208, 0.237] for the planning objective and 0.207 for a random ranking. The two scores pick the same top-scored candidate on 7.6\% of anchors. Counting each repeated scenario once leaves 454 scenarios, on which the observed score reaches 19.4\% and the planning objective 3.1\%. On Push-T the latent distance thus ranks above random once the predictor is bypassed, although the observed score still misranks 79.3\% of anchors. We do not report the measurement on PointMaze or Wall. On PointMaze our renders of a reached state differ from the stored frames by more than a tolerance fixed in advance, and on Wall the stored frames and states follow different transition rules. On MetaWorld we took the measurement on 100 of the 750 anchors, drawn at random with a seed fixed before any result was read. On those anchors the observed score reaches a Hit@1 of 55.0\% against the planning objective's 0.0\% and a random ranking's 5.9\%, and a top-1 regret of 0.041 [0.023, 0.065] against 0.276 [0.266, 0.285] and a random 0.258. When an anchor's candidates are scored against another anchor's goal, the observed score's Hit@1 changes little, to 58\% on MetaWorld and 18.3\% on Push-T. On Push-T every goal shows the block at the same target pose, and on MetaWorld the best candidate is the largest action along $+y$ on 619 of the 750 anchors, so on both tasks a score can rank these candidates well without using the goal. The observed score's high Hit@1 therefore shows that the predictor loses ranking information, not that the latent distance tracks a particular goal.

On MetaWorld the released checkpoint's top-scored candidate holds the gripper still on 491 of the 750 anchors and acts at random on 254 more. The anchor's best candidate is instead the largest action along $+y$ on 619 anchors and the expert's prefix, alone or with noise added, on 129 of the others, and the checkpoint picks neither on any anchor. The negative Spearman of Table~\ref{tab:scripted} therefore comes from a preference between kinds of candidate, not from a fine-grained misordering.

\subsection{Other encoders}
The encoders compared in Section~\ref{sec:ranking-encoders} are the image-pretrained DINOv2 ViT-S/14 \citep{oquab2023dinov2} and the video-pretrained V-JEPA 1 ViT-L/16 \citep{bardes2024vjepa}, V-JEPA 2 ViT-L/16 and ViT-g/16 \citep{assran2025vjepa2}, and V-JEPA 2.1 ViT-G/16 \citep{murlabadia2026vjepa21}. Each encoder is frozen and paired with a six-layer predictor of DINO-WM's architecture, trained per task without proprioceptive input and with a one-step rollout loss. Training runs 50 epochs at a constant learning rate of $5\times10^{-4}$, weight decay $10^{-7}$ and seed 234, on 224-pixel inputs for DINOv2 and 256-pixel inputs for the V-JEPA encoders, in batches of 16, or 8 for ViT-g and ViT-G. The ranking anchors are new simulator resets rather than frames from the training data, and each configuration is in the release. No validation set is used and no checkpoint is selected, so each audit scores the last checkpoint that training saves. In this comparison a Push-T candidate's true cost is one minus its highest dense reward, as in Table~\ref{tab:scripted}, and a MetaWorld candidate's is the Euclidean distance between the environment's observation vector at the end of the candidate and at the end of the expert rollout that defines the goal.

\begin{figure}[!t]
    \centering
    \includegraphics[width=\linewidth]{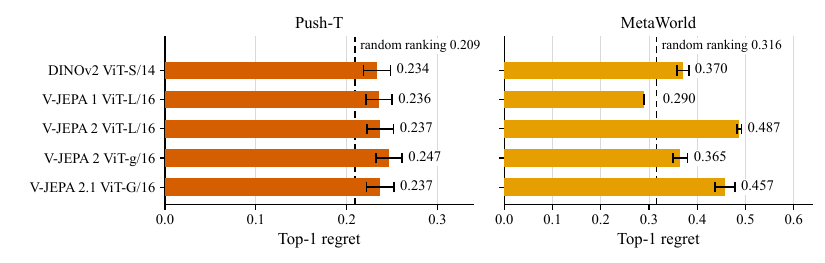}    \caption{\textbf{Top-1 regret on Push-T and MetaWorld for the five encoders of the encoder-swap comparison of Section~\ref{sec:ranking-encoders}.} This audit generates its own anchors, so its Push-T retention differs from Table~\ref{tab:scripted}'s: the bars cover 587 retained of 750 on Push-T and 750 on MetaWorld, with 95\% bootstrap intervals, and the number beside each bar is its regret. The dashed line is the regret of a random ranking of the thirty-two candidates. On Push-T every encoder has a higher regret than a random ranking, and on MetaWorld every encoder except V-JEPA 1 does.}
    \label{fig:encoders}
\end{figure}

\begin{figure}[!t]
    \centering
    \includegraphics[width=\linewidth]{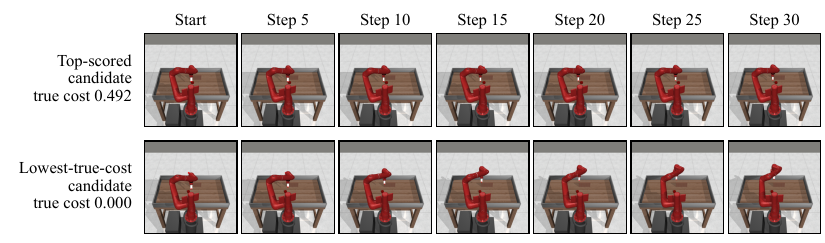}
    \caption{\textbf{Frames every five environment steps from executing two candidates of one anchor, for the V-JEPA 1 model of the encoder-swap comparison of Section~\ref{sec:ranking-encoders} on MetaWorld reach-wall.} The model is not a released checkpoint. The top row executes the candidate the model's planning objective scores best, and the bottom row the lowest-true-cost candidate, which is the prefix of the scripted expert's own actions. In the encoder-swap comparison a candidate's true cost is a distance between observations, not the distance between simulator states used elsewhere in the paper, so these true costs are not comparable with those of Table~\ref{tab:scripted}. The anchor shown has the largest top-1 regret of the 750, at 0.492. On 743 of the 750 anchors the model's top-scored candidate applies no action.}
    \label{fig:qual-metaworld}
\end{figure}

\begin{figure}[!t]
    \centering
    \includegraphics[width=\linewidth]{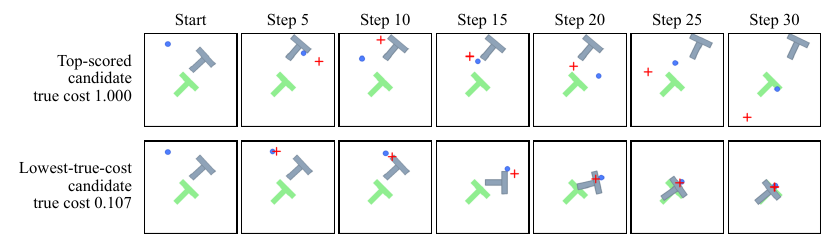}
    \caption{\textbf{Frames every five environment steps from executing two candidates of one anchor, for the V-JEPA 2 model of the encoder-swap comparison of Section~\ref{sec:ranking-encoders} on Push-T.} The model is not a released checkpoint. Of the anchor's thirty-two scripted candidates, the top row executes the one the model's planning objective scores best and the bottom row the one with the lowest true cost. A candidate's true cost is one minus its highest dense reward. The grey T is the block, the blue disc the pusher, the green T the target pose, and the red cross the pusher's start. The top-scored candidate pushes the block away from the target; the lowest-true-cost candidate pushes the block onto the target.}
    \label{fig:qual-pusht}
\end{figure}

All five models have a higher mean regret than a random ranking on Push-T, and four of five do on MetaWorld (Figure~\ref{fig:encoders}).
V-JEPA 1 is the exception on MetaWorld: its regret is lower than a random ranking's, but its top-scored candidate never has the lowest true cost. Each encoder has one training seed and one training recipe, so this is an observation about these checkpoints and not a statement about encoder families. Figures~\ref{fig:qual-metaworld} and~\ref{fig:qual-pusht} show one misranked anchor frame by frame for two of these models, on MetaWorld reach-wall and on Push-T.

\subsection{Models trained end to end}
For models trained end to end, we examine whether the latent distance favors actions from expert demonstrations.
Every model is scored on the same 500 anchors, each starting from an observation in a Push-T expert trajectory, and we compare the recorded continuation, the actions the expert took next, with alternative action sequences, using the later observation from that trajectory as the goal.
Among each anchor's 32 candidates, the recorded continuation has the lowest latent distance, including ties, at 69.6\% of anchors for released LeWM~\citep{maes2026leworldmodel}, 85.0\% for our 50-epoch LeWM, and 52.8\% for our Delta-JEPA~\citep{zhang2026deltajepa}, against about 5\% for a random ranking. Appendix~\ref{app:lewm-delta} gives the training of the two models we trained. A random ranking of the same thirty-two candidates puts the recorded continuation first, ties included, at 5.3\% of anchors for released LeWM, 5.8\% for our 50-epoch LeWM and 4.8\% for our Delta-JEPA; the three differ because a tie is a tie in the model's own scores, so each model induces its own tie structure on the shared candidates.

\section{Additional replanning results}
\label{app:replanning}

On PointMaze the three seeds give success rates of 85.4, 91.7 and 91.7\% with replanning, against 43.8, 68.8 and 72.9\% without it. On Push-T the three seeds give 45.8, 29.2 and 33.3\% against 10.4, 12.5 and 14.6\%. When the seeds are pooled, the paired difference is 27.8 points on PointMaze with a 95\% bootstrap interval of 19.4 to 36.1 points, and 23.6 points on Push-T with an interval of 16.7 to 30.6 points. The two-sided p-value from the exact paired test is below $10^{-8}$ on both tasks. Replanning after every planner action also runs the search six times per episode where planning once runs it once, so the two conditions differ in total search as well as in feedback. On PointMaze the extra search does not explain the difference: replanning at the reduced budget scores at most 1{,}920 candidates per episode and succeeds in 89.6\% of episodes, more than planning once at the larger budget of 9{,}000 candidates, which succeeds in 77.1\%. On Push-T planning once at the larger budget succeeds in 67.4\%, more than replanning at the reduced budget, so that comparison cannot tell the two causes apart. We therefore also run Push-T planning once with 384 rather than 64 candidates per CEM round, so that the single plan scores 1,920 candidates, the most that replanning scores in an episode, on the same 144 episodes. Planning once with that search succeeds in 70 of the 144 episodes (48.6\%), against 52 (36.1\%) with replanning and 18 (12.5\%) when planning once at the reduced budget. This condition succeeds more often than replanning in all three seeds, and 27 episodes succeed only under this condition against 9 only with replanning (exact paired test, p = 0.0039). On Push-T, spending the same search on a single plan therefore matches and exceeds the gain from replanning at this budget, so there we find no benefit from the new observations that replanning uses.

\section{Statistical tests}
\label{app:stats}

The NR intervals of Tables~\ref{tab:nr-full} and~\ref{tab:controls-full} are 95\% percentile bootstrap intervals over episodes with 2{,}000 resamples, and the success rates of Table~\ref{tab:success-full} carry 95\% Wilson intervals. Unless otherwise stated, the two scorers in every comparison are measured on the same anchors and the same fixed candidate sets. Apart from the episode-level intervals quoted for Figure~\ref{fig:teaser}, which are 95\% Wilson intervals, every confidence interval in Appendix~\ref{app:ranking} and in the closed-loop runs of Appendix~\ref{app:replanning} is a 95\% percentile bootstrap, over anchors at 10{,}000 resamples for the scripted-candidate audit and over paired episodes at 20{,}000 for the runs. Whether a checkpoint's Hit@1 falls below a random ranking's is tested one-sided against the exact distribution of the number of anchors a random ranking gets right, in which each anchor counts with the probability that a random pick ties for its lowest true cost. The probability of a count as low as the checkpoint's is $3.3\times10^{-5}$ on Push-T, with 15 of 618 anchors right against 36.0 expected, and $7.3\times10^{-21}$ on MetaWorld, with 0 of 750 against 44.9. On Push-T the top-1 regret, whose interval overlaps a random ranking's, cannot be told apart from it, so the below-random finding there rests on Hit@1. On each task 200 of the 750 anchors repeat an earlier anchor's seed and true costs, which leaves 454 distinct scenarios among Push-T's 618 retained anchors and 550 on MetaWorld, and the encoder comparison's MetaWorld anchors repeat in the same way. Counting each scenario once, the tail probability is $3.6\times10^{-3}$ on Push-T, with a Hit@1 of 3.1\% against 5.9\%, and $1.7\times10^{-15}$ on MetaWorld, with 0 of 550 against 33.0 expected. The top-1 regret is then 0.219 [0.201, 0.236] against a random 0.206 on Push-T and 0.281 [0.275, 0.288] against 0.257 on MetaWorld. Paired binary outcomes are compared with the exact paired test, McNemar's test on the discordant pairs, whether the outcomes are two conditions' episode successes or two scorers' top-1 correctness. Every result the paper calls significant survives a Holm correction for as many as twenty tests, more than the paper reports.

\section{Attempts to fix the ranking}
\label{app:fixing}

The two scorers of Section~\ref{sec:discussion} are two attempts that ranked fixed candidates better without raising planning success, and they do not exhaust the options. We give here how they were built, the candidates and episodes on which they were measured, and the per-condition numbers the main text leaves out.

\begin{figure}[!t]
    \centering
    \includegraphics[width=\linewidth]{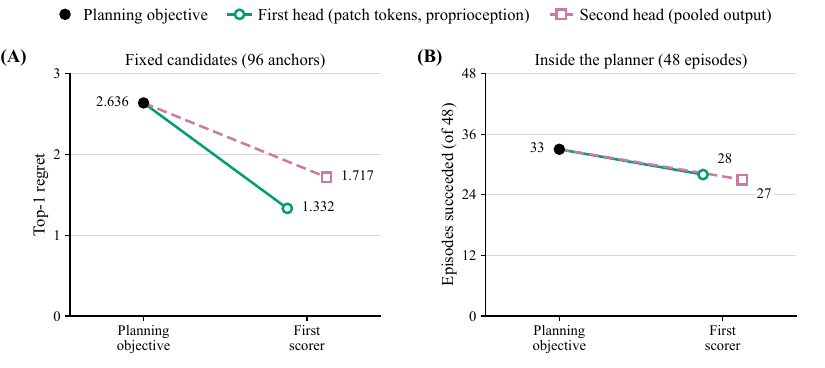}
    \caption{\textbf{The first scorer against the planning objective, on fixed candidates and inside the planner.} (A) Top-1 regret on 96 development anchors with 64 candidates each, for the planning objective and for the first scorer's two heads. (B) Episodes that succeed, out of 48, under the planner's own selection rule and when the planner's candidates are reranked with each head. The second head's line is dashed to stay visible where the two lines nearly coincide. Better ranking on fixed candidates does not carry into the planner.}
    \label{fig:first-scorer}
\end{figure}

For the first scorer, our re-implementation of the reachability cost of \citet{li2026trm}, we trained two heads over the frozen encoder's outputs, each an ensemble of three seeds. One head reads the encoder's patch features and the proprioceptive state separately before combining them, and the other reads the encoder's pooled output. Both heads were trained with a Smooth-L1 loss on how many frames apart two frames of one released trajectory are. We measured the first scorer twice, once on a set of candidates fixed in advance and once inside the planner (Figure~\ref{fig:first-scorer}). The fixed set holds 96 development anchors with 64 candidates each. On that set a candidate's true cost is its full-state cost, the distance from its terminal state to the goal's full state, velocity included. On that set the top-1 regret falls from 2.636 for the planning objective to 1.332 for the first head and 1.717 for the second. The first head's improvement has a 95\% interval of [0.970, 1.633], and when the 96 anchors are split into six groups the first head has a lower regret than the planning objective in all six groups. Inside the planner the first scorer was run on 48 development anchors, one episode each, one plan per episode, with five CEM rounds of 64 candidates and eight elites. The planner's own selection rule succeeds in 33 of those episodes. Reranking with the first head succeeds in 28 and with the second head in 27. Using the first head as the search objective succeeds in 21, and a shuffled-label control succeeds in 5. Over the first 24 anchors the planner's own selection rule succeeds in 17 and reranking with the first head in 13. In 28 episodes across the two first-head variants, the planner's own selection rule succeeds and the first scorer fails. In 25 of those 28, once velocity is counted, the plan the first scorer picked ends closer to the goal than the one the planner's own selection rule picked, and in position alone it ends no closer.

On Push-T the second scorer predicts from the frozen encoder's outputs how much of the environment's fixed target region each candidate will cover. The second scorer's runs judge success differently from the rest of the paper. An episode counts as a success when the T-shaped block covers 95\% of the fixed target region at any executed step. The second scorer was trained on 48 anchors with 64 candidates each. On those anchors no candidate reaches the coverage that counts as a success at any step, even when every candidate is executed first and the best is kept. Executing every candidate and keeping the best yields a rise in the mean highest dense reward from 0.229 under the planner's own selection rule to 0.357. That rise is the most the second scorer could recover. The second scorer recovers 51\% of the rise from 0.236 under the planning objective's own top candidate on held-out data, and 48\% of the corresponding rise on a further 48 anchors. A third set of 48 anchors was executed once with the second scorer and once without it. On that set the highest dense reward rises by 0.068 with a confidence interval of [0.027, 0.113] when the planner plans once, and by 0.028 with a confidence interval of [0.002, 0.055] when the planner replans after every action. In both cases the success count is 0 of 48 with the second scorer and 0 of 48 without it.

\section{Scope of the recommendation}
\label{app:scope}

The measurements behind the recommendation cover two released models on four simulated tasks. The real-robot dataset in JEPA-WM's release has no simulator in which to compute a true cost, so the measurements of this paper cover the four simulated tasks alone. Every candidate of an episode has to be executed from exactly the same state, which needs an environment that can be reset, so a real robot is out of scope here. A world model on a real robot also faces sources of error that simulation lacks. \citet{assran2025vjepa2} report one such source, a camera placement that leaves the direction of an action ill-defined for the model. Whether the ranking persists on a real robot, and how well the score picks there, remains untested. A scorer that picks the right plan would also save planner calls. In our setting, replanning after every action calls the planner six times per episode and planning once calls the planner once, so the planner is called 83\% less often. The saving would be large, because planning with V-JEPA 2-AC on a real robot takes about 16 seconds per action \citep{assran2025vjepa2}.

\end{document}